\pdfoutput=1
\documentclass{article}
\usepackage{hyperref}
\usepackage{spconf,amsmath,graphicx}
\usepackage{hyperref}
\usepackage{array}
\usepackage{multirow,amsfonts,cite,url}

\title{Dynamic Multi-scale context aggregation for conversational Aspect-based Sentiment Quadruple Analysis}
\name{
    Yuqing Li$^{1,2}$ \quad
    Wenyuan Zhang$^{1,2}$ \quad 
    Binbin Li $^{1*}$\thanks{*Corresponding author} \quad 
    Siyu Jia$^1$ \quad 
    Zisen Qi$^1$ \quad 
    Xingbang Tan$^1$}
\address{
    $^1$ Institute of Information Engineering, Chinese Academy of Sciences   \\
    $^2$ School of Cyber Security, University of Chinese Academy of Sciences  
}
\begin{document}
%
\maketitle
\begin{abstract}
Conversational aspect-based sentiment quadruple analysis (DiaASQ) aims to extract the quadruple of target-aspect-opinion-sentiment within a dialogue .
In DiaASQ,  a quadruple's elements often cross multiple utterances. This situation complicates the extraction process, emphasizing the need for an adequate understanding of conversational context and interactions.
However, existing work independently encodes each utterance,
thereby struggling to capture long-range conversational context and overlooking the deep inter-utterance dependencies. 
In this work, we propose a novel Dynamic Multi-scale Context Aggregation network (DMCA) to address the challenges. Specifically, we first utilize dialogue structure to generate multi-scale utterance windows for capturing rich contextual information. After that, we design a Dynamic Hierarchical Aggregation module (DHA) to integrate progressive cues between them. In addition, we form a multi-stage loss strategy to improve model performance and generalization ability. Extensive experimental results show that the DMCA model outperforms baselines significantly and achieves state-of-the-art performance\footnote{The code is available at \url{https://github.com/qdCassie-Li/DMCA}}.
\end{abstract}
\section{Introduction}
\label{sec:intro}
\begin{figure}[!t]
\vspace{10pt}
  \centering
  \includegraphics[width=0.89\columnwidth]{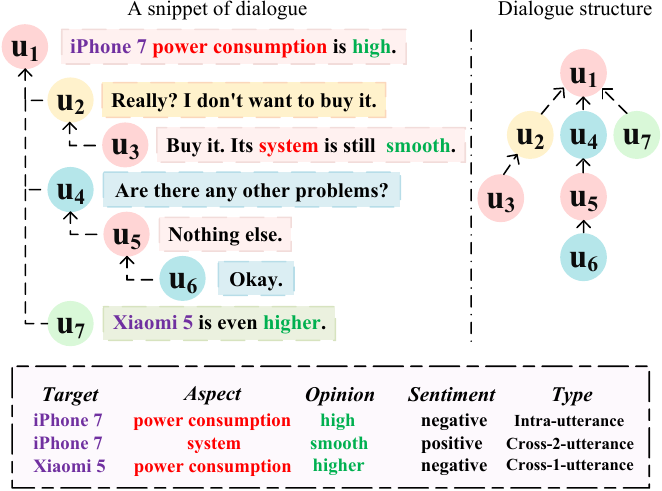}
  \caption{
   Conversational aspect-based sentiment quadruple analysis task with its corresponding outputs.
  Utterances are represented as nodes in the tree, with the color indicating the speaker, and the structure presents reply relationships.
  }
  \label{fig:example}
  \vspace{-11pt}
\end{figure}

In recent years, sentiment analysis of reviews has gained increasing attention.
Broad applications include stance detection\cite{DBLP:conf/emnlp/AugensteinRVB16}\cite{DBLP:conf/coling/SunWZZ18}, document-level \cite{DBLP:conf/coling/LyuFG20}\cite{DBLP:conf/acl/TangQL15} and aspect-based \cite{DBLP:conf/icassp/KeXWLY22}\cite{DBLP:conf/ijcai/0001LLWLJ22}\cite{DBLP:conf/emnlp/WuYZFDX20} sentiment analysis. Recent research\cite{DBLP:conf/acl/Li0LWZWLLLCJ23} has broadened the scope of sentiment analysis to incorporate dialogue-level reviews, called the conversational aspect-based sentiment quadruple analysis (DiaASQ), which reflects more realistic dialogue-driven user review scenarios.
DiaASQ aims to predict the quads $\{(\textbf{t},\textbf{a},\textbf{o},\textbf{s})\}$ from a dialogue. As shown in Fig.~\ref{fig:example},
multiple speakers express their reviews around several \underline{\textbf{t}}argets (iPhone 7 and Xiaomi 5). They emphasize different \underline{\textbf{a}}spects (power consumption and system), while expressing their respective \underline{\textbf{o}}pinions (high and smooth). The \underline{\textbf{s}}entiment is determined based on the opinion of the target.

In contrast to sentiment tuples extraction focuses on independent sentence~\cite{DBLP:conf/acl/XuCB20} \cite{DBLP:conf/emnlp/ZhangD0YBL21}, 
DiaASQ expands extraction perspective to the dialogue.
Uniquely, a quadruple might span across several utterances,
so a comprehensive understanding of the dialogue and the context of utterances is crucial.
Despite previous research~\cite{DBLP:conf/acl/Li0LWZWLLLCJ23} efforts to mitigate this limitation through attention mechanisms and positional encoding techniques, 
it still faces challenges in capturing the semantic interactions and rich contextual information in multi-turn dialogues. 
Relevant works \cite{DBLP:conf/nips/ZaheerGDAAOPRWY20}\cite{DBLP:conf/icml/ZhangGSL0DC21} have proposed methods for PLMs to adapt to longer inputs, but they mainly focus on attention mechanisms\cite{DBLP:conf/nips/VaswaniSPUJGKP17} or network architectures, rather than capturing critical information from dialogues.
Fixed-size sliding window methods are commonly used for processing long dialogues\cite{DBLP:conf/aaai/ZhongLX0022}\cite{DBLP:conf/emnlp/XiaSZXLYWHZL22}, but they overlook the benefits of multi-scale windows which can capture richer context.

\begin{figure*}[t]
  \centering
  \includegraphics[width=0.9\linewidth]{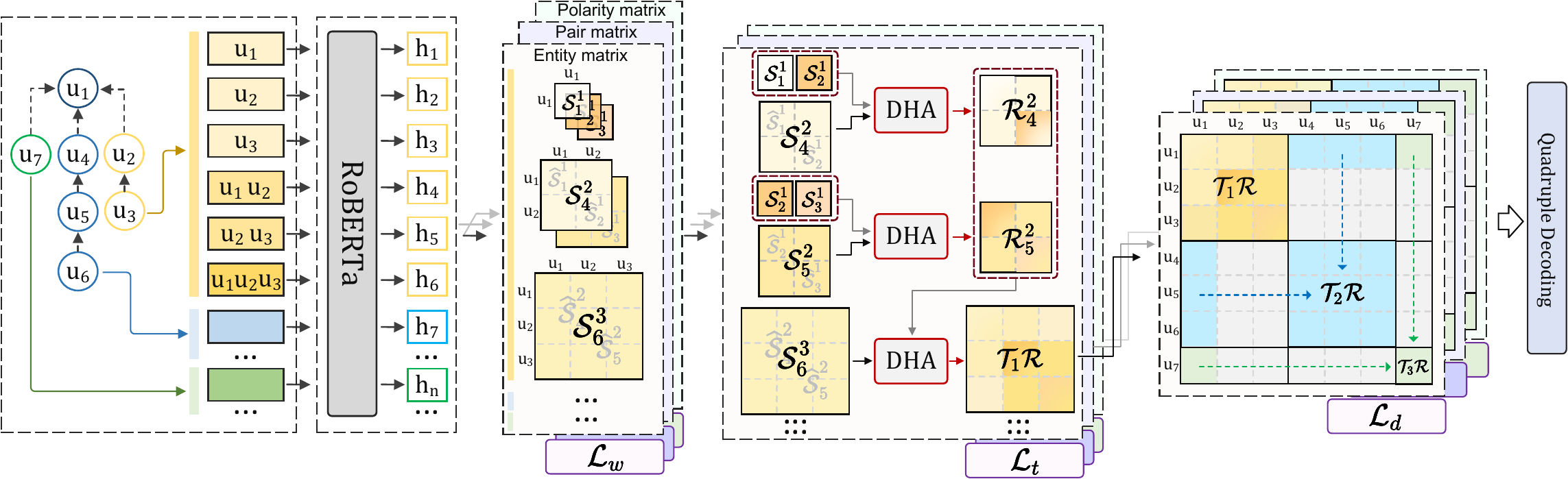}
  \caption{The overall framework of our DMCA model. The model consists of two key components: 1) a flexible sliding window scheme that captures conversational context at multiple scales and granularities, and 2) a Dynamic Hierarchical Aggregation (DHA) module along with
a multi-stage loss strategy that hierarchically aggregates the logits of multi-scale windows. Note: The third dimension of the logits has been omitted from the matrix for clearer visualization.
 }
  \label{fig:overall}
\end{figure*}

In this paper, we propose a novel \textbf{D}ynamic \textbf{M}ulti-scale \textbf{C}ontext \textbf{A}ggregation network (DMCA) for DiaASQ, as shown in Fig.~\ref{fig:overall}.
\textbf{Firstly}, we employ a flexible sliding window scheme to create variable-sized utterance windows.
This approach facilitates the comprehensive capturing of dialogue context, ranging from a single utterance to broader spans.
\textbf{Secondly}, we introduce a \textbf{D}ynamic \textbf{H}ierarchical \textbf{A}ggregation (DHA) module.
The goal of DHA is to enhance dialogue quadruple prediction by aggregating the output logits from multi-scale windows, eliminating the necessity for intricate network designs.
Specifically, 
DHA hierarchically uses logits from smaller windows as a basis to aggregate and update the logits of larger windows that encompass these smaller windows. This process continues until aggregated logits are obtained at the dialogue level. 
\textbf{Furthermore}, we introduce multi-stage losses to jointly optimize different levels of aggregation, including window-level, thread-level, and dialogue-level.
We conduct extensive experiments on two public benchmark datasets, and the results prove that DMCA significantly outperforms comparative methods. 

The main contributions are summarized as follows: 
1) We introduce the DMCA network to improve the extraction of dialogue quadruples by utilizing multi-scale context.
2) Without relying on complex network architectures, we design the Dynamic Hierarchical Aggregation module (DHA) along with multi-stage losses to optimize the decision-making process.
3) Extensive experiments show that the DMCA significantly outperforms state-of-the-art methods.
\section{Methodology}
\label{sec:format}
\subsection{Problem Definition and Preliminaries}
A dialogue is denoted as $\{(u_i, s_i,r_i)\}|_{i=1}^{|D|}$, where utterance $u_i$ is uttered by the speaker $s_i$ and is in response to $u_{r_i}$. $|D|$ denotes the total number of utterances.
Based on the aforementioned input, the goal of the task is to predict all the sentiment quadruples $Q\!=\!\{(\textbf{t},\textbf{a},\textbf{o},\textbf{s})\}$, where each quadruple contains: target($t$), aspect($a$), opinion($o$), and sentiment polarity($s$). Here, sentiment polarity $\in \{pos, neg, other\}$.

\noindent\textbf{\textit{Tagging schema.}}
To transform the extraction of dialogue quadruples into a unified grid tagging task,
we follow the tagging strategy of previous work\cite{DBLP:conf/acl/Li0LWZWLLLCJ23}. 
Specifically, the dialogue quadruple extraction task is broken down into three joint tasks: detection of \textit{entity boundaries (ent)}, \textit{entity relation pairs (pair)}, and \textit{sentiment polarity (pol)}.
In the \textit{entity boundaries} detection phase, `tgt', `asp', and `opi' are used to respectively represent the head and tail relations of the target, aspect, and opinion items between any word pairs within the window. 
In the \textit{entity relation pair} detection phase, the labels `h2h' and `t2t' are used to align the head and tail markers between two types of entities. 
For instance, `iPhone' (target-head) and `power' (aspect-head) are connected by `h2h', while `7' (target-tail) and `consumption' (aspect-tail) are connected by `t2t'. 
Sentiment labels $\{pos, neg, other\}$ are obtained in \textit{sentiment polarity} detection.
By combining the results derived from these three tasks, we can efficiently extract the complete dialogue quadruples.
\subsection{DMCA Model}
\label{sec:init-prior}
\subsubsection{Multi-scale context windows generation}
\label{sec:multi}
A set of utterances within a dialogue with a complete reply relationship is defined as a thread\cite{DBLP:conf/acl/Li0LWZWLLLCJ23}. Studies\cite{DBLP:conf/sigir/VedulaCR23}\cite{DBLP:journals/coling/ElsnerC10} have delved into the independence between dialogue threads. To effectively capture the rich context, we use a sliding window method to construct multi-scale windows for each thread.
Firstly, we analyze the dialogue structure using the reply records $\{r_i\}_{i=1}^{|D|}$, treating each dialogue branch as an independent thread. This gives rise to a collection of threads \( T=\{T_t\}_{t=1}^{|T|} \), where \( |T| \) represents the number of threads and each thread \( T_t =\{u_1,u_j,\cdots,u_{j+\ell_t-1}\} \) consists of \( \ell_t \) utterances. For each thread, we use a flexible sliding window schema to generate continuous subsets from the thread. We denote these subsets as windows, represented by \( W^t=\{W^t_w\}_{w=1}^{|W^t|} \). The size of these windows varies from 1 to \( \ell_t \). Therefore, for each thread, the total number of windows \( |W^t| \) is determined by the formula \( |W^t| = \frac{1}{2}(\ell_t^2 + \ell_t) \). We have verified that all generated windows meet the input requirements of the PLM.
Consider the illustration in Fig.~\ref{fig:example}, where \( T_1 = \{u_1,u_2,u_3\} \). It can produce 6 distinct windows: \{$u_1$\}, \{$u_2$\}, \{$u_3$\}, \{$u_1,u_2$\}, \{$u_2,u_3$\} and \{$u_1,u_2,u_3$\}.

Secondly, we encode windows to obtain representations:
\begin{equation}
\small
\mathbf{H}^t_w  = [\textbf{h}_{[CLS]},\textbf{h}_1,\cdots\textbf {h}_{N_w},\textbf{h}_{[SEP]}] = \text{Encoder}\left(W_{w}^t\right),
\end{equation}
\begin{equation}
\small
W_{w}^t = \{[CLS];u_1;u_j;\cdots u_{j+k-1};[SEP]\}.
\end{equation}
We use RoBERTa \cite{DBLP:journals/corr/abs-1907-11692} as Encoder.
$\textbf{H}^t_w\in \mathbb{R}^{N_w\times D_h}$ denotes the representation of $W_w^t$ . $N_w$ is the number of tokens in the window and $D_h$ is hidden size.

Subsequently, we obtain the output logits of the word pair matrix, denoted as 
$ \mathcal{S}_w = \{s_{ij} \,|\, i, j \in [1, N_w]\}$. 
Additionally, we introduce a window-level cross-entropy loss $\mathcal{L}_w$ to supervise predictions at a more granular level for each window:
\begin{equation}
\small
    \widetilde{\textbf{h}}_i  =  \widetilde{\textbf{W}}\textbf{h}_i + \widetilde{\textbf{b}},
\end{equation}
\begin{equation}
\small
s_{ij}=(\widetilde{\textbf{h}}_i)^T\widetilde{\textbf{h}}_j,
\end{equation}
\begin{equation}
\small
p_{ij} = \text{Softmax}(s_{ij}),
\end{equation}
\begin{equation}
\small
\mathcal{L}_{w} = -\sum_{w=1}^{|W|}\sum_{i=1}^{N_w}\sum_{j=1}^{N_w}y_{ij}\log(p_{ij}),
\label{eq:loss}
\end{equation}
where 
$s_{ij}\in \mathbb{R}^K$, $K$ represents the predefined number of categories in the decoding table and $y_{ij}$ represents the truth label. $\widetilde{\textbf{W}}$ and $\widetilde{\textbf{b}}$ are trainable parameters.
\subsubsection{Dynamic Hierarchical Aggregation module}
Windows of different scales capture distinct information: smaller windows focus on local details, while larger ones emphasize contextual understanding.
We introduce the Dynamic Hierarchical Aggregation (DHA) module to aggregate predicted logits from these windows, avoiding the need for designing complex network architectures.
This aggregation process is categorized into thread-level and dialogue-level.
\\
\textbf{\textit{Thread-level Aggregation.}}
The predicted logits of all windows within the $t$-th thread are denoted as $\mathcal{S} = \{\mathcal{S}_{i} \mid u_i \in T_t\}$. Adding a superscript $l$ indicates the number of utterances comprising the window. 
DHA utilizes the $\mathcal{S}^l_{i}$ from the small window $W^t_i$ to aggregate and augment the $\mathcal{S}^{l+1}_{j}$ of larger overlapping window $W^t_j$, while ensuring that $W^t_i\subseteq W^t_j$. 
Specifically, we extract logits corresponding to \(W^t_i\) from \(\mathcal{S}^{l+1}_{j}\) to form \(\hat{\mathcal{S}}^{l}_{i}\). To enhance the predictions in the larger window, we select logits among \({\mathcal{R}}^{l}_{i}\), \(\hat{\mathcal{S}}^l_i\), and \({\mathcal{R}}^l_{i}+\hat{\mathcal{S}}^l_i\) based on the principle of minimizing cross-entropy. 
These selected logits are then aggregated using a weighted summation approach.
This process updates $\mathcal{S}^{l+1}_{j}$ to ${\mathcal{R}}^{l+1}_{j}$. 
The definition of this dynamic aggregation process is as follows:
\begin{equation}
\small
{\mathcal{R}}^{l+1}_{j} = \mathcal{S}^{l+1}_{j} \oplus \alpha \cdot \mathcal{F}^l_{i},
\end{equation}
\begin{equation}
\small
\mathcal{F}^l_{i} = \mathop{\arg\min}_{x \in \mathcal{X}^{l}_{i} }{CrossEntropy}(x,y),
\label{eq:label}
\end{equation}
\begin{equation}
\small
 {\mathcal{X}}^{l}_{i} = \{{\mathcal{R}}^{l}_{i},\hat{\mathcal{S}}^l_i,  {\mathcal{R}}^l_{i}+\hat{\mathcal{S}}^l_i\},
\end{equation}
where $\oplus$ denotes the broadcast addition.  $ \alpha $ is a predefined parameter. 
Padding($\cdot$) implies zero-padding.  $y$ denotes corresponding truth labels.
The initial value for $ {\mathcal{R}}^1_{i}$ is set as $\mathcal{S}^1_{i}$. 

Through the  dynamic hierarchical process, we obtain the aggregated thread-level logits as:
${\mathcal{T}_t\mathcal{R}} = {\mathcal{R}}^{\ell_t}_{|W^t|} $. 
The thread-level loss \( \mathcal{L}_{t} \) is calculated in a manner analogous to Eq.~\ref{eq:loss}. Notably, DHA is only used during the training phase since it requires label information (Eq.~\ref{eq:label}). For validation and test, we adopt Static Hierarchical Aggregation (SHA). The SHA approach hierarchically aggregates the logits of overlapping windows through a direct sum operation. SHA is defined as:
\begin{equation}
\small
{\mathcal{R}}^{l+1}_{j} = \mathcal{S}^{l+1}_{j} \oplus {\mathcal{R}}^l_{i}
\label{eq:2}
\end{equation}
\textbf{\textit{Dialogue-level Aggregation.}}
After the aggregation process at the thread level, we obtain refined logits for each thread. Since these threads overlap only at the root utterance \( u_1 \), we utilize the SHA method to derive final dialogue-level logits  \( \mathcal{DR}\in \mathbb{R}^{N\times N\times K} \) and subsequently obtain $\mathcal{L}_d$. 
\begin{equation}
\small
   Padding({\mathcal{T}_{|T|}\mathcal{R}})
    \mathcal{DR} = {\mathcal{T}_1\mathcal{R}}\oplus\cdots\oplus {\mathcal{T}_{|T|}\mathcal{R}}
\end{equation}
\begin{equation}
\small
   \mathcal{L}_{d} = -\frac{1}{N^2}\sum_{i=1}^{N}\sum_{j=1}^{N}y_{ij}\log(p_{ij})
\end{equation}
where N denotes the number of tokens in the dialogue.
\label{sec:dha}
\subsubsection{Training}
During the training process, we jointly incorporate three distinct stages of loss: $\mathcal{L}_w$, $\mathcal{L}_t$, and $\mathcal{L}_d$.
These losses are employed to minimize errors at different aggregation stages. For each task $\psi$, the loss can be calculated as follows:
\begin{equation}
\small
\mathcal{L}^{\psi}= \mathcal{L}_d^{\psi} + \eta \mathcal{L}_t^{\psi} + \zeta \mathcal{L}_w^{\psi}
\end{equation}
where $\psi \in \{ent,pair,pol\}$, $\eta$ and $\zeta$ are predefined weights.

The final objective function is determined by the sum of the loss for the three tasks:
\begin{equation}
\small
\mathcal{L}= \mathcal{L}^{ent} + \mathcal{L}^{pair} + \mathcal{L}^{pol}
\end{equation}
\section{Experiments}
\label{sec:typestyle}
\begin{table*}[h]
\setlength\tabcolsep{3.3pt}
\footnotesize
\centering
  \begin{tabular}{l|ccc|ccc|cc||ccc|ccc|cc}
  \hline
  \multicolumn{1}{c|}{\multirow{3}{*}{\small \text{Model}}} & \multicolumn{8}{c||}{\text{ZH-dataset}} & \multicolumn{8}{c}{\text{EN-dataset}} 
  \\ 
  \cline{2-17}
& \multicolumn{3}{c|}{\text{Entity detection}} & \multicolumn{3}{c|}{\text{Pair detection}} & \multicolumn{2}{c||}{\text{Quads extraction}} & \multicolumn{3}{c|}{\text{Entity detection}} & \multicolumn{3}{c|}{\text{Pair detection}} & \multicolumn{2}{c}{\text{Quads extraction}} \\
   & \textit{T} &\textit{A} & \textit{O} & \textit{T-A} & \textit{T-O} & \textit{A-O} & {{micro-F1}} & {{iden-F1}} 
  & \textit{T} &\textit{A} & \textit{O} & \textit{T-A} & \textit{T-O} & \textit{A-O} & {{micro-F1}} & {{iden-F1}}     \\
  \hline
     Extract-Classify & 91.11 & 75.24 & 50.06 & 32.47 & 26.78 & 18.90 & 8.81 & 9.25  & 88.31 & 71.71 & 47.90 & 34.31 & 20.94 & 19.21 & 11.59 & 12.80  \\ 
     SpERT &  90.69 & 76.81 & 54.06 & 38.05 & 31.28 & 21.89 & 13.00 & 14.19 &  87.82 & 74.65 & 54.17 & 28.33 & 21.39 & 23.64 & 13.07 & 13.38  \\
     ParaPhrase & / & / & / & 37.81 & 34.32 & 27.76 & 23.27 & 27.98 & / & / & / & 37.22 & 32.19 & 30.78 & 24.54 & 26.76  \\
    Span-ASTE & / & / & / & 44.13 & 34.46 & 32.21  & 27.42  & 30.85 & / & / & / & 42.19 & 30.44 &  45.90  & 26.99  & 28.34 \\
    DiaASQ & 90.23 & {76.94} & 59.35 &  48.61 &  43.31 &  45.44 & 34.94 & 37.51 &  \bf{88.62} &  \bf{74.71} &  60.22 &  47.91 &  45.58 & 44.27 &  33.31 &  36.80 \\
    \hline
    \textbf{Ours(DMCA)}  
    & \bf 92.03    & 	\bf{77.07}   & 	\bf{60.27} & 	\textbf{56.88} & 	\textbf{51.70} & 	\bf{52.80} & 	\textbf{42.68}   & 	\textbf{45.36}   
    & {88.11}   &   {73.95}  &  \bf{63.47}  &  \bf{53.08}  &  \textbf{50.99} & 	\textbf{52.40}   & 	\textbf{37.96} &   \textbf{41.00}\\
  \hline
  \end{tabular}    
  \caption{We report the micro-F1 scores for all tasks and the additional identification F1 (iden-F1)\cite{DBLP:conf/acl/BarnesKOOV20} scores for quads extraction. Here, T-A-O stands for Target-Asepct-Opinion, respectively.
  }
  \label{tab:main}
\end{table*}
\begin{table}[h]
\setlength\tabcolsep{2pt}
\small
\centering
\begin{tabular}{p{2cm}|>{\centering\arraybackslash}p{1.5cm}>{\centering\arraybackslash}p{1.5cm}|>{\centering\arraybackslash}p{1.5cm}>{\centering\arraybackslash}p{1.5cm}}
\hline
\multicolumn{1}{c|}{\multirow{2}{*}{\textbf{Methods}}} & \multicolumn{2}{c|}{\multirow{1}{*}{\textbf{ZH}}}  & \multicolumn{2}{c}{\multirow{1}{*}{\textbf{EN}} }
\\ 
\cline{2-5}
  & \multirow{1}{*}{{micro-F1}} & \multirow{1}{*}{{iden-F1}} & \multirow{1}{*}{{micro-F1}} & \multirow{1}{*}{{iden-F1}}  \\ \hline
  {\ \ \  }\textbf{DHA}	 & \textbf{42.68}   & 	\textbf{45.36}  &  \textbf{37.96} &	\textbf{41.00}  \\ \hline
  {\ \ \  }SHA(Eq.~\ref{eq:2})	    & 42.31         & 	44.92 & 	37.73 & 	39.91 \\  
  {\ \ \  }Concat       & 41.24  & 	43.50   & 	34.75 & 	37.31 \\
\hline
\end{tabular} 
\caption{Results against different aggregation methods.  `Concat' denotes the direct concatenation of logits from the largest window across all threads.
} \label{tab:aggregation}
\end{table}
\begin{table}[!h]
\setlength\tabcolsep{4pt}
\small
\centering
\begin{tabular}{p{2cm} c c c}
  \hline
  \textbf{Methods} & \textbf{Intra} & \textbf{Inter} & \textbf{Overall}\\ 
  \hline
  DMCA & \textbf{46.23} & \textbf{32.73} & \textbf{42.68} \\ 
  \hspace{5pt} - w/o $\mathcal{L}_w$ & 46.05($\downarrow$0.18) & 31.78($\downarrow$0.95) & 42.43($\downarrow$0.25) \\
  \hspace{5pt} - w/o $\mathcal{L}_t$ & 45.10($\downarrow$1.13) & 27.74($\downarrow$4.99) & 40.57($\downarrow$2.11) \\
  \hspace{5pt} - w/o $\mathcal{L}_d$ & 45.17($\downarrow$1.06) & 30.94($\downarrow$1.79) & 41.51($\downarrow$1.17) \\  
  \hline
\end{tabular}      
\caption{Ablation results of DMCA. We report the micro-F1 score for the ZH dataset. `Inter' denotes the score of cross-utterance quadruple extraction.}
\label{tab:loss}
\end{table}

\subsection{Tasks and Datasets}
We conduct experiments on two datasets: the Chinese dataset \textbf{ZH} \cite{DBLP:conf/acl/Li0LWZWLLLCJ23} and the English dataset \textbf{EN} \cite{DBLP:conf/acl/Li0LWZWLLLCJ23}. Both datasets contain speaker and reply-record information for each conversation utterance. Each dataset consists of 1000 dialogues related to electronic product reviews, with an average of 7 utterances and 5 speakers per dialogue. Specifically, the Chinese dataset contains 5,742 quadruples, while the English dataset contains 5,514 quadruples. 
About 22\% of the quadruples in both datasets are cross-utterance.
\subsection{Comparison Methods}
\textbf{\textit{Baseline.}} Following the comparison in \cite{DBLP:conf/acl/Li0LWZWLLLCJ23}, we consider several powerful performance models closely tied to the task as baselines. These models include {ExtractClassify} \cite{DBLP:conf/acl/CaiXY20}, {SpERT} \cite{DBLP:conf/ecai/EbertsU20}, 
 {Span-ASTE} \cite{DBLP:conf/acl/XuCB20}, {ParaPhrase}\cite{DBLP:conf/emnlp/ZhangD0YBL21}, and {DiaASQ}\cite{DBLP:conf/acl/Li0LWZWLLLCJ23}.
\\
\textbf{\textit{Implementation Details.}}
To encode \textbf{ZH} and \textbf{EN} datasets, we take the Chinese-Roberta-wwm-base \cite{DBLP:journals/taslp/CuiCLQY21} and RoBERTa-Large \cite{DBLP:journals/corr/abs-1907-11692}, respectively.  Each training process contains 25 epochs.
The parameters for both the DHA module (\(\alpha\)) and the loss (\(\eta\) and \(\zeta\)) are initialized to 1 by default.
For the tasks $\psi \in \{ent, pair, pol\}$, the values of $K$ is $\{6, 4, 3\}$. We use micro F1 and identification F1\cite{DBLP:conf/acl/BarnesKOOV20} as the evaluation metrics.
\subsection{Results and Analysis}
\begin{figure}[!t]
  \centering
  \includegraphics[width=0.84\columnwidth]{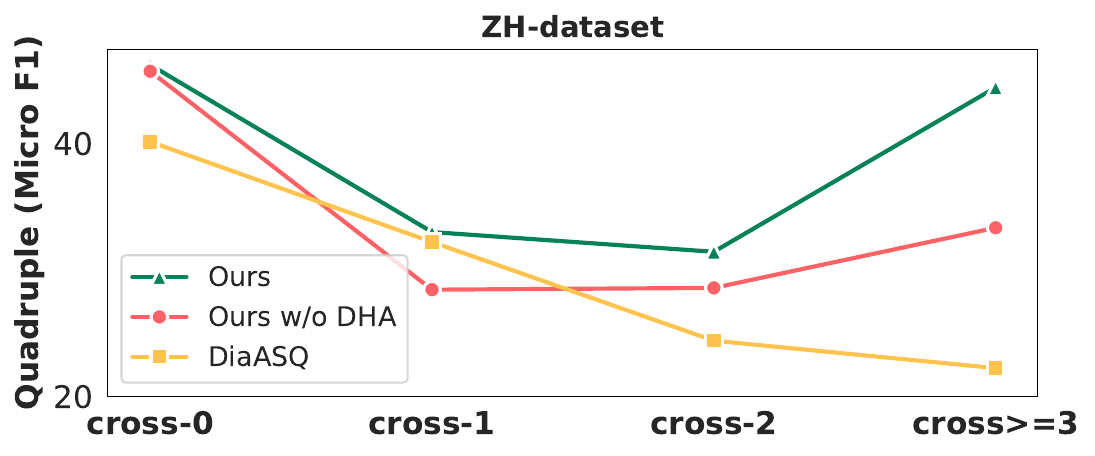}
  \caption{
  Results of cross-utterance quadruples. `cross-0' indicates elements of the quadruple contained in one utterance.
  }
  \label{fig:cross2}
  \vspace{-10pt}
\end{figure}

\textbf{\textit{Overall Results.}}
The overall results are shown in Table~\ref{tab:main}. 
Our model outperforms the previous best baseline in almost all tasks and datasets. 
Notably, on ZH dataset, 
our model surpasses the previous state-of-the-art by an impressive 7.7\%.\\
\textbf{\textit{Cross-Utterance Results.}}
To further demonstrate the effectiveness of DMCA model in addressing cross-utterance quadruple extraction, we conduct a detailed analysis and comparison of the cross-utterance results, as shown in Fig.~\ref{fig:cross2}. Our approach outperforms previous model in all cross-utterance counts, especially achieving high performance when cross $\ge 3$. This indicates that DMCA model is more effective in handling extraction problems in multi-turn dialogues.
\subsection{Ablation}
We conduct experiments to assess the impact of the DHA module and the three distinct stage loss functions.
As shown in Table~\ref{tab:aggregation}, the DHA method, which considers the credibility of predictions from multi-scale windows, achieves the highest performance. 
Without the dynamic weighted aggregation, the performance of the SHA method diminishes.
When we remove the aggregation module, the results significantly decline on both datasets, highlighting the success of our DHA.
Moreover, as depicted in Table~\ref{tab:loss}, removing any stage of the loss function results in a decrease in performance, particularly for the problem of cross-utterance extraction. This further demonstrates the effectiveness of the multi-stage losses.
\section{Conclusion}
In this paper, we propose a novel DMCA network for conversational aspect-based sentiment quadruple analysis. To address the challenges of encoding long dialogues and extracting cross-utterance quadruples, we construct multi-scale utterance windows to capture rich dialogue context. 
We also design a DHA module and multi-stage loss stategy to enhance the decision-making logits from these multi-scale windows.
Experimental results on two datasets demonstrate the superiority of our DMCA over the state-of-the-art methods.
\vfill\pagebreak

\label{sec:refs}
\bibliographystyle{IEEEbib}
\bibliography{strings,refs}

\end{document}